\documentclass[twoside,11pt]{article}

\usepackage[abbrvbib,preprint]{jmlr2e}

\usepackage{graphicx}

\usepackage{amsmath,amssymb}
\DeclareMathOperator{\E}{\mathbb{E}}

\firstpageno{1}

\title{A Survey on Self-supervised Pre-training for Sequential Transfer Learning in Neural Networks}

\author{\name Huanru Henry Mao \email hhmao@eng.ucsd.edu \\
       \addr Computer Science and Engineering\\
       University of California, San Diego}

\begin{document}

\maketitle

\begin{abstract}
Deep neural networks are typically trained under a supervised learning framework where a model learns a single task using labeled data.
Instead of relying solely on labeled data, practitioners can harness unlabeled or related data to improve model performance, which is often more accessible and ubiquitous.
Self-supervised pre-training for transfer learning is becoming an increasingly popular technique to improve state-of-the-art results using unlabeled data.
It involves first pre-training a model on a large amount of unlabeled data, then adapting the model to target tasks of interest.
In this review, we survey self-supervised learning methods and their applications within the sequential transfer learning framework.
We provide an overview of the taxonomy for self-supervised learning and transfer learning, and highlight some prominent methods for designing pre-training tasks across different domains.
Finally, we discuss recent trends and suggest areas for future investigation.
\end{abstract}

\section{Introduction}
Deep learning has led to significant improvements in state-of-the-art performance across many domains \citep{lecun2015deep} and has become the dominant approach in building intelligent systems over the last decade.
Traditionally, deep networks are trained under a supervised learning framework where a model is trained \textit{tabula rasa} (from scratch) to optimize the performance on a single task with the hopes of generalizing to unseen test examples.
A task is typically provided as a set of labeled data with the assumption that the training and test set are drawn from the same underlying distribution.
While effective when labeled data is abundant, the paradigm of learning a single task in isolation is limited when human-annotated data is lacking for tasks of interest, leading to poor model generalization \citep{DBLP:journals/tkde/PanY10}.

In contrast to the supervised learning framework, humans are able to learn priors about our environment without labels and adapt our knowledge to new tasks with only a few examples \citep{DBLP:conf/iclr/DubeyAPEG18}.
For instance, learning how to play piano can help us learn music fundamentals, which, subsequently, makes learning how to play violin easier.
When an infant learns how to recognize faces, they can apply this knowledge to recognize other objects \citep{wallis1999learning}.
Ideally, a similar approach could be applied to machine learning.
Instead of relying solely on labeled data, practitioners can leverage unlabeled or related data, which is often more accessible and ubiquitous.
Knowledge from a large corpus of unlabeled data can be extracted and transferred to improve performance on a target task where labeled data is either limited or unavailable.

There is a large amount of literature on unsupervised and transfer learning.
In this paper, we focus on surveying \textit{self-supervised learning} methods for sequential transfer learning.
Self-supervised learning is a type of unsupervised learning where a model is trained on labels that are automatically derived from the data itself without human annotation \citep{DBLP:journals/jmlr/ErhanBCMVB10,DBLP:journals/neco/HintonOT06}.
Self-supervised learning methods enable a model to learn useful knowledge about an unlabeled dataset by learning useful representations and parameters.
Transfer learning focuses on how to transfer or adapt this learned knowledge from a \textit{source task} to a \textit{target task}  \citep{DBLP:journals/tkde/PanY10}.
Specifically, we focus on a specific type of transfer learning called sequential transfer learning \citep{ruder2019neural} which adopts a ``pre-train then fine-tune" paradigm.
Self-supervised learning and transfer learning are two complementary research areas that, together, enable us to harness a source task with a large amount of unlabeled examples and transfer the learned knowledge to a target task of interest.
These methods have grown in popularity due to their success and scalability in improving state-of-the-art results across domains.
Finding useful self-supervised learning algorithms and transfer learning methods are areas of active investigation.

Compared to other surveys that focus primarily on either computer vision \citep{schmarje2020survey,DBLP:journals/corr/abs-1902-06162} or natural language processing (NLP) \citep{ruder2019neural}, we provide a broad review of self-supervised learning across domains in computer vision, natural language and audio/speech.
This can, hopefully, provide a birds eye view of self-supervised research in deep learning and highlight areas for further investigation.

We first provide a background overview of self-supervised pre-training and transfer learning in section \ref{sec:tl} and \ref{sec:ssl}.
We then review self-supervised learning methods organized under the following categories: bottleneck-based methods (sec. \ref{sec:bottleneck}) and prediction-based methods (sec. \ref{sec:prediction}).
Bottleneck-based methods drive learning by imposing an information bottleneck through a model's architecture.
Prediction-based methods learn by asking a model to predict or generate relevant data with respect to the input.
Finally, we provide a discussion of research trends and frontiers for future work in section \ref{sec:frontiers}.

\section{Transfer Learning}
\label{sec:tl}
We provide a more formal definition of transfer learning, following the definitions from \citet{DBLP:journals/tkde/PanY10,ruder2019neural}.
Transfer learning is a collection of techniques that focus on adapting knowledge between tasks and involves two concepts: a domain and a task.
A domain $\mathcal{D}=\{\mathcal{X},P(X)\}$ has a feature space $\mathcal{X}$ and a marginal distribution $P(X)$ over the feature space where $X=\{x_1,\ldots,x_n\} \in \mathcal{X}$.
For image classification, $\mathcal{X}$ is the space of all images, $x_i$ corresponds to some image and $X$ is a sample of images used for training.

A \textit{task} $\mathcal{T}$ is defined with respect to some domain $\mathcal{D}$ and consists of a label space $\mathcal{Y}$, a prior distribution $P(Y)$ and a learned conditional probability distribution $P(Y|X)$.
$P(Y|X)$ is typically learned from a training set of $\{x_i,y_i\}$ where $ x_i \in X, y_i \in \mathcal{Y}$.
For image classification, $\mathcal{Y}$ is the set of possible image classes.

The aim of transfer learning is to learn the target task $\mathcal{T}_t$
using knowledge learned from a source task $\mathcal{T}_s$.
Specifically, we want to learn the target conditional probability distribution $P_t(Y_t|X_t)$ in $\mathcal{D}_t$ from information learned from $\mathcal{T}_s$ and $\mathcal{D}_s$ where $\mathcal{D}_s \neq \mathcal{D}_t$ and/or $\mathcal{T}_s \neq \mathcal{T}_t$.

\subsection{Transfer Learning Scenarios}
\begin{figure}
  \centering
  \includegraphics[width=0.8\linewidth]{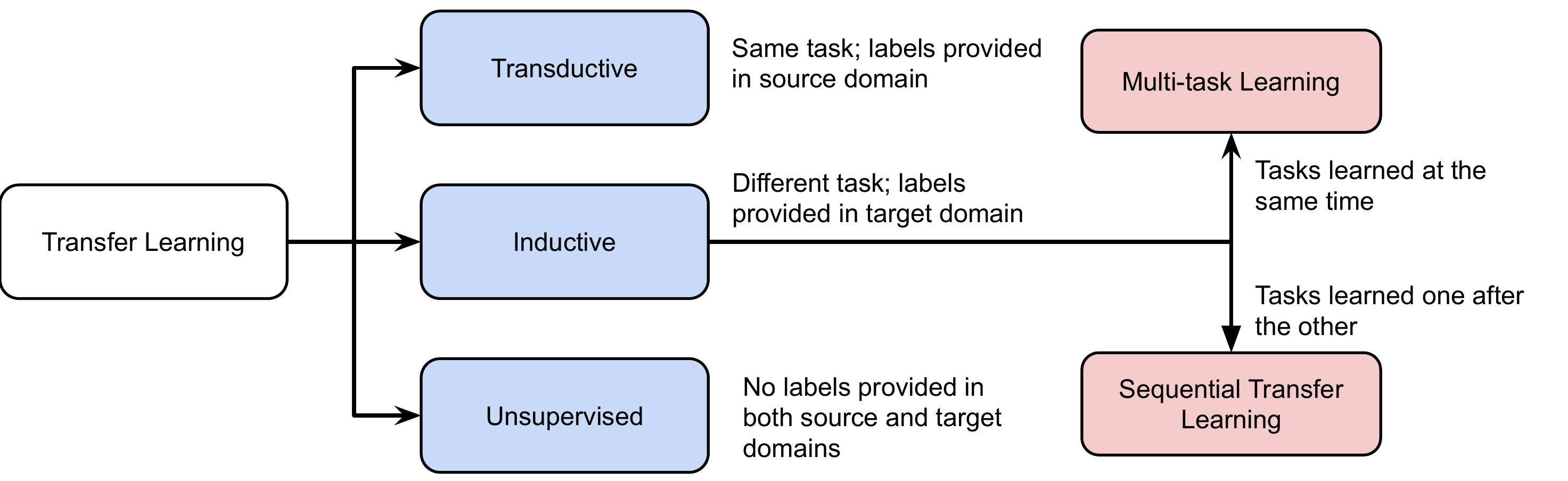}
  \caption{Transfer learning scenarios based on \citet{ruder2019neural}.}
  \label{fig:tl_tax}
\end{figure}

Following the taxonomy from \citet{DBLP:journals/tkde/PanY10}, we can categorize transfer learning into three broad settings depending on the tasks in the source and target domain (Figure \ref{fig:tl_tax}).
When source and target tasks are the same $\mathcal{T}_s = \mathcal{T}_t$ with labels only available in the source domain, we call it transductive transfer learning.
A specific example of transductive transfer learning is domain adaptation \citep{wang2018deep}, where a model could be trained on the source task of predicting sentiment on Amazon reviews and needs to be adapted to predict sentiment in news.
When the tasks are different and labeled data is provided in the target domain $\mathcal{T}_s \neq \mathcal{T}_t$, we refer to this as inductive transfer learning.
An example of inductive transfer learning is training a model on (optionally labeled) data to classify many images of natural scenery, then adapting the model to classify images of cats.
When no labels are provided in either case, we refer to this setting as unsupervised transfer learning.

\citet{ruder2019neural} further refines inductive transfer learning into two subcategories: multi-task learning \citep{caruana1997multitask} and sequential transfer learning.
In multi-task learning, tasks $\mathcal{T}_s$ and $\mathcal{T}_t$ are learned simultaneously, typically through the joint optimization of multiple objective functions.
In sequential transfer learning, $\mathcal{T}_s$ is first learned, then the downstream task $\mathcal{T}_t$ is learned.
The first stage is often called \textit{pre-training} while the second stage of learning is often called \textit{fine-tuning} in the context of neural networks.
The primary difference between these two types of transfer learning is \textit{when} the target task is learned.
More generalized schemes can define a multi-tasking schedule that interpolates between learning the source and target task \citep{liang2017deep} or contain multiple source tasks \citep{mao2019improving}.

Sequential transfer learning is our primary focus in this paper.
Sequential transfer learning is more popular in practice as it is simple to set up a two-phase training pipeline and easy to distribute pre-trained models without needing to disclose the pre-training dataset.
Most of the self-supervised learning techniques we review can be categorized under sequential transfer learning.

\paragraph{Related Areas}
There are other related research areas to transfer learning that are beyond the scope of this paper, that we briefly mention here.
Lifelong learning \citep{parisi2019continual} can be seen as form of sequential transfer learning of many tasks, with the additional goal of learning without forgetting previous tasks (e.g.,~catastrophic forgetting).
Few shot learning \citep{wang2019generalizing} focuses on the general problem of learning with few labels and is achievable in certain extents with transfer learning.
Meta learning \citep{vilalta2002perspective} focuses on algorithms that enable us to learn how to learn and can be considered a form of transfer learning where meta-knowledge is transferred to task-specific knowledge.

\subsection{Why Does Sequential Transfer Learning Work?}
In order to understand the success of sequential transfer learning it is useful to consider some theoretical arguments as to why it works.

\paragraph{Multi-tasking Perspective}
We briefly summarize an analysis from \citet{DBLP:journals/corr/Ruder17a} on why multi-task learning is beneficial since it is highly related to sequential transfer learning.
Multi-task learning \citep{caruana1997multitask} has been shown to serve as a form of regularization as it reduces the Rademacher complexity of the model (the ability to fit random noise) \citep{sogaard2016deep}.
It biases the model to prefer representations that other tasks would likely prefer \citep{baxter2000model} and allows the model to learn a task better through \textit{hints} from another task \citep{abu1990learning}.
While multi-task and sequential transfer learning are not strictly the same, it is useful to consider these related effects especially when hybrid sequential and multi-task transfer learning approaches are used.

\paragraph{Regularization}
To understand why sequential transfer learning works, we summarize an early work that provides useful insights in unsupervised pre-training. \citet{DBLP:journals/jmlr/ErhanBCMVB10} analyzed a special case of unsupervised pre-training applied to deep belief networks \citep{DBLP:journals/neco/HintonOT06}, but the arguments presented there are more broadly applicable to sequential transfer learning.
\citet{DBLP:journals/jmlr/ErhanBCMVB10} hypothesizes that pre-training serves as a form of implicit regularization through parameter initialization by constraining the minima that the supervised objective can optimize to.
Pre-training restricts learning to a subset of the parameter space bound by a basin of attraction achievable through fine-tuning the supervised target task.
This hypothesis is supported experimentally by observing the training dynamics of MNIST filters \citep{DBLP:journals/jmlr/ErhanBCMVB10}.
Recent work has also shown some evidence to suggest that fine-tuning pre-trained language models does not deviate from the pre-trained weights significantly \citep{DBLP:journals/corr/abs-2005-07683}.
In other words, the final weights are mostly pre-determined by pre-training, especially if the pre-training task dominates the total training time of the sequential transfer learning process.

\paragraph{Inducing Priors}
Treating sequential transfer learning as simply a form of regularization underestimates its benefits.
Similar to multi-task learning, sequential transfer learning also induces a prior on the model.
Practitioners who use similar source and target tasks encode a prior on what knowledge is likely useful, thus the effects are akin to selecting good neural architectures and better hyperparameters.

\paragraph{Implicit Meta Learning}
Another perspective we can consider is that pre-training, when given an appropriate and sufficiently large source task, can perform implicit meta learning \citep{DBLP:journals/corr/abs-2005-14165}.
This provides a similar effect as meta-learning algorithms such as MAML \citep{finn2017model} that explicitly aim to learn an initialization that easily adapts to various problems.








\section{Self-supervised Learning}
\label{sec:ssl}
Unsupervised learning is a family of approaches that learn from data without any supervision.
A particular form of unsupervised learning of growing interest is self-supervised learning.
The terms unsupervised and self-supervised have been, historically, used interchangeably in the literature, but recent work has preferred the term self-supervised learning for its specificity.
In this review, we refer to self-supervised learning as any unsupervised learning approach that can be easily reduced into a supervised problem by generating labels.
Thus, self-supervised learning can reap the advancements and breakthroughs from supervised learning.
Self-supervised learning still requires labels, but it is unsupervised in the sense that these labels are derived from the data itself rather than annotated by humans.

Early work in self-supervised pre-training for deep neural networks aimed to effectively train stacked auto-encoders \citep{bengio2007greedy} and deep belief networks (DBN) \citep{DBLP:journals/neco/HintonOT06} without labels.
These techniques train deep networks one layer at a time in a greedy fashion in order to circumvent poor local minima that prevented successful end-to-end gradient descent \citep{40d5d7fd62cb44ba934a8a75d4b2b076}.
Once trained, the neural network is \textit{fine-tuned}, where the model with pre-trained weights switches from unsupervised learning to supervised learning objective of the target task.
This can lead to improved performance on the target task as opposed to simply learning the target task from scratch.
In the last decade, greedy layer-wise unsupervised learning has fallen out of fashion in favor of end-to-end learning where an entire deep network is trained in one operation.
This shift is partly due to the architectural innovations \citep{he2015deep}, normalization \citep{ioffe2015batch} and better activation functions \citep{nair2010rectified} that enable training of very deep networks \citep{bachlechner2020rezero} while avoiding local minima.

In contrast to classic work on greedy self-supervised learning \citep{ackley1985learning,demers1993non}, modern approaches focus on end-to-end learning. 
Self-supervised learning constructs a pre-training or ``pretext'' task that is used to extract knowledge from unlabeled data \citep{DBLP:journals/corr/abs-1902-06162}.
After training a model on the pretext task, it can then be adapted to the target task through transfer learning.
Pre-training tasks come in many forms.
They usually involve transforming or imputing the input data with the goal of forcing the model to predict missing parts of the data or through introducing some information bottleneck, which we will review in later sections.

\paragraph{Downstream Tasks}
Self-supervised learning has been used to transfer knowledge into a variety of target tasks.
In this review, we do not focus on any specific downstream task since we are primarily concerned with pre-training methods.
Instead, we briefly highlight some common tasks used to benchmark self-supervised learning algorithms.
For computer vision, image classification is typically the downstream task of interest for self-supervised learning of still images \citep{chen2020simple} and action recognition benchmarks are used to evaluate video self-supervised learning methods \citep{srivastava2015unsupervised}.
For natural language, a popular benchmark, GLUE \citep{wang2018glue}, has been used to test self-supervised learning approaches on a bag of tasks including natural language inference, sentiment analysis and paraphrase identification.
For speech, automatic speech recognition, phoneme identification and speaker identification are downstream tasks of interest \citep{chi2020audio}.

\paragraph{Information Content in Learning}
It is useful to consider the amount of information content that can be derived from different learning frameworks \citep{ruder2019neural}.
Yann LeCun has referred to the hierarchy of information given to learning algorithms metaphorically as a ``cake'': Reinforcement learning gets the cherry on top (a single scalar value per episode, \citep{williams1992simple}); supervised learning the frosting (10-10k bits per sample); and unsupervised and self-supervised learning is the foundation of the cake (millions of bits per example depending on the domain). 
Hence, in many cases, self-supervised learning can provide significantly more information per example for learning.

\subsection{Generative verses Discriminative Learning}
A critical decision when designing pre-training schemes is to consider whether we want to perform generative or discriminative learning.
In this section we outline the differences between the two approaches. 
These approaches are somewhat orthogonal to the choice of pre-training tasks and either option can be used for a given task.

\subsubsection{Generative Approaches}
Generative approaches for self-supervised learning involve the process of producing all or parts of the training data as part of the model's output \citep{DBLP:journals/corr/abs-1902-06162}.
For instance, we can take a frame in a video and ask the model to generate future frames \citep{srivastava2015unsupervised}.
The labels, in this case, are typically in the feature space of the training data.
Generative approaches have the advantage that the output is qualitatively interpretable as we can inspect samples from the model.
In addition, generative models have other applications beyond self-supervised learning \citep{goodfellow2014generative}.
The drawback of generative learning is that it requires learning how to produce every single detail in the input feature space, which could be a substantial amount of dedicated computation and modeling resources.
For example, generating an image requires predicting every single pixel in the output space of the model and the process of decoding an image is not necessarily helpful for transfer learning to downstream tasks.

For continuous domain applications such as images or raw audio, generation is challenging when there are multiple ``correct" answers (e.g., predicting the future audio frames spoken), sometimes leading to the model predicting the mean of all futures (which qualitatively results in blurry predictions).
To avoid generating the average prediction, researchers have adopted alternative generative techniques using adversarial learning (GANs)  \citep{goodfellow2014generative}, which can lead to sharper generations.
For a detailed survey on GANs, we refer the reader to \citet{jabbar2020survey}.

\subsubsection{Discriminative Approaches}
On a high level, discriminative approaches for self-supervised learning involve the process of determining positive samples from negative samples.
When labels are provided, as in supervised learning, this is simply called classification.
Discriminative approaches eschew the challenge of generation by asking the model to simply differentiate between pairs of input samples.
In self-supervised learning, a common interpretation of discriminative learning without labels is mutual information maximization \citep{hjelm2018learning}.

The mutual information (MI) \citep{bell1995information} of two random variables $X, Y$ measures the reduction in uncertainty of one variable when the other is observed.
For instance, knowing that the background of an image contains grass $x$ can make us less uncertain about the location $y$ in which the image was photographed.
For the purpose of self-supervised learning, it may be desirable to maximize the mutual information between certain features of the data \citep{hjelm2018learning}.

More formally, mutual information is defined as:
\begin{equation}
    I(X,Y) = \E_{p(X,Y)} \left[ \log \frac{p(x,y)}{p(x)p(y)} \right]
\end{equation}
It is intractable to compute $I$ and sample-based estimators that maximize lower bounds on MI are used in practice.
The most commonly used lower bound that has been shown to work well is Information Noise Contrastive Estimation (InfoNCE) \citep{DBLP:journals/corr/abs-1807-03748}.
InfoNCE is a probabilistic contrastive loss \citep{chopra2005learning} that tries to separate positive examples from negative examples.
Following the formulation and notation in \citet{kong2019mutual}, the InfoNCE lower bound is defined as:
\begin{equation}
   I(A,B) \geq \E_{p(A,B)} \left[ f_\theta(a,b) - \E_{q(\tilde B)} \left[ \log \sum_{\tilde b\in \tilde B}{\exp f_\theta (a,\tilde b)} \right] \right] + \log | \tilde B |,
   \label{eq:infonce}
\end{equation}
where $a$ and $b$ are the positive example pairs, $\tilde B$ is a set of samples drawn from some proposal distribution $q(\tilde B)$, and $f_\theta \in \mathbb{R}$ is a learned comparison function with parameters $\theta$.
$\tilde B$ contains positive samples $b$ and $|\tilde B| - 1$ negative samples.
There are many ways to construct $f_\theta$.
For instance, we can construct it as the dot product of features produced by two identical encoders, commonly known as Siamese Networks \citep{hadsell2006dimensionality}.

In practice, training $f_\theta$ involves sampling a pair of positive samples and $|\tilde B| - 1$ negative samples, then minimizing the cross entropy loss of the positive example over all samples.
This is equivalent, in expectation, to maximizing Eq. \ref{eq:infonce}.

Contrastive learning can be used in self-supervised learning by trying to predict certain samples from negative samples, such as predicting future audio frames against random frames or image patches within the same images against random patches (details in section \ref{sec:prediction}).
This works well for various continuous domain tasks as shown in \citet{DBLP:journals/corr/abs-1807-03748}.
A challenge with contrastive learning is choosing proposal distribution $q(\tilde B)$, which determines how negative samples are selected.
Having a large number of negative samples can be helpful in certain domains \citep{he2020momentum}.

For discrete domain tasks such as natural language, \citet{kong2019mutual} show that language modeling and generation tasks that maximize cross entropy loss also maximizes InfoNCE.
Indeed, cross entropy loss is a special case of InfoNCE when $\tilde B = B$.
For instance, language modeling predicts the next token by comparing against all possible tokens in the model's vocabulary.
This is equivalent to performing a ``negative sampling" scheme where all possible outputs are sampled at all times.

Mutual information maximization alone is insufficient for learning good representations as suggested in \citet{hjelm2018learning} and demonstrated empirically in \citet{DBLP:conf/iclr/TschannenDRGL20}.
Instead, good representations also depend on the choice of architecture, task and parametrization of the MI estimators.

\section{Architectural Bottleneck Methods}
\label{sec:bottleneck}
We categorize self-supervised learning approaches that primarily rely on an information bottleneck induced through a model's architecture as bottleneck-based methods.
Bottleneck-based methods attempt to learn a low dimensional or constrained representation of the data typically by learning to reconstruct the input data \citep{demers1993non}.
Bottleneck-based methods are sometimes categorized in the literature as unsupervised rather than self-supervised learning.

By learning a constrained representation, a model must discard irrelevant information and retain useful information.
A direct application of bottleneck-based methods is in their learned representations, which can be used as feature extractors for downstream target tasks \citep{fleming1990categorization,cottrell1990face,cottrell1991empath}.
Alternatively, the model's weights (usually the encoder) can also be transferred to target tasks via fine-tuning or learned jointly in a multi-task setting.
We present a summary of bottleneck-based methods in this section.

\subsection{Dimensionality Reduction}
We first briefly review classical approaches to dimensionality reduction.
The most well known technique for dimensionality reduction in machine learning is principal component analysis (PCA) \citep{wold1987principal}.
Given a dataset of $d$-dimensional vectors represented as a matrix $X$, PCA aims to find a low dimensional representation of the data by eliminating correlations between variables.
In practice, PCA can be solved by using singular value decomposition.
The low dimensional feature can then be used as input to other machine learning algorithms.

\subsubsection{Latent Semantic Analysis}
In natural language, inputs are often sequences of discrete tokens, which can be represented as one-hot vectors over some vocabulary.
It is useful to extract low-dimensional representation of words, known as \textit{word embeddings}, because one-hot vectors are large in dimensionality and do not contain semantic meaning of the words they represent.

Latent semantic analysis (LSA) \citep{DBLP:journals/jasis/DeerwesterDLFH90} is a classic technique used to extract low dimensional distributed representations of words based on the co-occurrence of words within a document context.
First, a word-document matrix $A$ is constructed by counting the occurrence of words that appear in each document.
Then, we apply dimensionality reduction on $A$ using singular value decomposition to factorize it into the product of three matrices. 
\begin{equation}
A = U \Sigma V^T
\end{equation}
Word embeddings $E$ of dimension $d$ can be extracted by truncating matrices $U$ and $\Sigma$ by retaining only the top $d$ rows, resulting in $U_d$ and $\Sigma_d$. Then, word embeddings can be computed by the product:
\begin{equation}
E = U_d \Sigma_d
\end{equation}
LSA can be considered as PCA applied to matrix $A$.

\subsection{Deep Autoencoders}
Approaches such as PCA and LSA extract low dimensional representations of data, but are linear approaches.
Deep autoencoders \citep{demers1993non,hinton2006reducing} with non-linearity are more expressive approaches that can extract better low-dimensional features from data.
In the most high level definition, deep autoencoders can be formulated as two neural networks that contain an encoder $\texttt{enc}$ and decoder $\texttt{dec}$.
The encoder produces a latent representation from input $x$,
\begin{equation}
    h = \texttt{enc}(x).
\end{equation}
While the decoder reconstructs the input $x$ from the latent representation $h$,
\begin{equation}
    \hat{x} = \texttt{dec}(h).
\end{equation}
Autoencoders are trained to minimize the reconstruction error between $x$ and $\hat{x}$.
Thus, autoencoders are considered as generative approaches.

There are a variety of methods to impose an information bottleneck in the autoencoder and we summarize some prominent approaches in the follow subsections.
Once an autoencoder has been trained, depending on the task, the decoder may be discarded and the encoder can be transferred to downstream tasks.
We also highlight how some of these techniques have been applied for fine-tuning.

\subsubsection{Compression-based Autoencoders}
In order to learn a non-trivial mapping, the dimensionality of $h$ is typically constrained to be less than the dimensionality of $x$, thus the model must learn what information to keep.
Early work \citep{munro1989image} demonstrated that this bottleneck, after quantization, can learn effective image compressors.

Compression-based autoencoders have been successfully applied to natural language for self-supervised learning.
In \citet{DBLP:conf/nips/DaiL15}, a sequence-to-sequence autoencoder is trained to take an input sequence $x$, produce a single latent vector $h$, which is then used to generate the original sequence $x$.
The authors demonstrated improvements on sentiment analysis tasks through pre-training.

\subsubsection{Sparse Autoencoders}
Alternatively, $h$ can be \textit{overcomplete} (dimensionality of $h$ is greater than dimensionality of $x$) but constrained by other means \citep{bengio2013representation}.
One popular constraint is by imposing a sparsity prior on the latent representation typically by minimizing the L1 loss of $h$.
Sparsity prior has been motivated biologically by the human visual cortex \citep{olshausen1997sparse}.
\citet{ranzato2008sparse} suggests that this acts as a soft way of restricting ``the volume of the input space over which the energy surface can take a low value".

\citet{makhzani2013k} introduced k-sparse autoencoders, where the latent representation is constrained to only have top $k$ largest activations active and the rest are set to zero.
They demonstrated experimentally that k-sparse autoencoders can be used as a pre-training step, then fine-tuned on image classification tasks for improved performance.
One drawback of this technique is that it could lead to dead hidden units, which can be addressed by scheduling the sparsity level.

\subsubsection{Variational Autoencoders}
Another method to impose a constraint on the latent variables is by treating the latent variable as a stochastic variable, as introduced in variational autoencoders (VAE) \citep{kingma2013auto}.
VAEs impose a regularization term in addition to the reconstruction loss to minimize the KL divergence between the latent representation and a prior distribution, typically a multivariate Gaussian.
One issue with VAE approaches in sequence models is the \textit{posterior collapse} problem, where the latent variable is completely ignored when more powerful sequence decoders are used \citep{roberts2018hierarchical}.
This can be mitigated by using hierarchical decoders.
We have found that, generally, Gaussian VAEs have been less explored for the purpose of transfer learning.

The prior distribution of VAEs could also be categorical.
Vector quantized variational autoencoders (VQ-VAE) \citep{van2017neural} and probabilistic variants \citep{sonderby2017continuous} are a type of autoencoder with a discrete bottleneck.
The learned discrete bottleneck provides several advantages such as enabling latent discrete modeling.

Discrete latent autoencoders have been used in speech for unsupervised phoneme discovery.
For example, \citet{eloff2019unsupervised} trained autoencoders to quantize speech and found that the learned discrete codes can be used for speech synthesis.
Quantizing speech is a reasonable prior, since spoken words in raw wave forms often have corresponding discrete phonemes that represent them.
Discrete latent models have also been combined with prediction-based methods for self-supervised speech recognition \citep{baevski2019vq}.
In these scenarios, the extracted discrete latent code can be further processed using NLP pre-training techniques, such as BERT \citep{devlin2018bert}, to learn even better representations.
\citet{dhariwal2020jukebox} extended this line of work to learn discrete latent codes for music generation.

\subsection{Other Approaches}
Autoencoders are not the only way to impose bottleneck learning.
\citet{wu2018unsupervised} demonstrate that one can perform unsupervised learning by simply treating each data point as its own class, and to maximally scatter all data points onto a 128 embedding space using contrastive learning.
By trying to compress the entire dataset into a low dimensional space similar inputs must cluster together.
They show that this method can lead to competitive results on ImageNet when compared to other self-supervised techniques.
This is similar to the bottleneck-based approaches in autoencoders except learning is performed using a contrastive loss without a decoder.

In \citet{donahue2016adversarial} the authors train a bidirectional GAN (BiGAN) for self-supervised representation learning.
GANs typically employ a generator and a discriminator, which learns a latent to data space mapping.
For self-supervised learning, we ideally want a data to latent mapping for downstream tasks (e.g.~image classification).
Thus, the authors propose a bidirectional GAN learning framework where an encoder is also learned jointly with a generator and discriminator.
This model is not explicitly an autoencoder, but the adversarial constraint forces the encoder to invert the generator.
The authors show that BiGAN is closely related to autoencoders with an $l_0$ loss.

\subsection{Limitations}
As mentioned in previous sections, bottleneck-based methods have shown success in various domains, especially when realized as autoencoders.
However, bottleneck approaches have generally been found to be inferior to prediction-based methods \citep{DBLP:conf/cvpr/ZhangIE17} and current state-of-the-art techniques are mostly prediction-based methods.
This may stem from the fact that bottleneck approaches need to trade off between information content and representational capacity.
Critics claim that autoencoders are an unsupervised learning approach, which, by definition, cannot be tailored to downstream tasks without additional priors \citep{rasmus2015semi}.
That is not to say that bottleneck approaches cannot be combined with prediction-based methods for improved performance and, indeed, many prediction-based methods have built upon bottleneck-based approaches.


\section{Prediction-Based Methods}
\label{sec:prediction}
Prediction-based methods aim to learn useful representations of data through learning a relevant predictive task such as asking the model to predict the missing parts of an input given its related context.
These techniques range from asking a model to predict the future given the present, predict missing patches of an image or missing words in a sentence.
Intuitively, prediction forces a model to learn relationships between the global and local parts of the data. 
In this section, we review methods for continuous and discrete domains separately since they tend to have different methods.

\subsection{Pre-training for Continuous Domains}
In this section, we focus on self-supervised learning methods for continuous domain tasks such as vision and speech.
An overarching theme among these approaches is to create self-supervised tasks that learn high level features while discarding low level information and noise.
Here, we summarize various commonly used pre-training tasks.

\subsubsection{Spatial Prediction}
\label{sec:spatial_pred}
Spatial prediction aims to learn representations by removing patches of an image and predicting the masked patches.
When posed as a generative task, this technique is also known as image in-painting.
In Context Encoders \citep{DBLP:conf/cvpr/PathakKDDE16}, the authors train a convolutional neural network (CNN) autoencoder by blanking-out the center patch of an input image and ask the model to generate contents within the missing square.
This is similar to a denoising autoencoder \citep{DBLP:conf/icml/VincentLBM08}, but differs in that the input mask is a contiguous block instead of random noise, and only the masked segments are predicted.
An issue raised by the paper is that pixel-level prediction creates blurry in-paintings, since L2 loss encourages learning the average of all possible completions.
Using an adversarial loss can mitigate this issue.

Alternative spatial masking approaches perform self-supervised learning by using a discriminative loss where the ground truth patch must be correctly identified from negative samples.
\citet{hjelm2018learning} proposes to maximize mutual information between local and global features of an image.
This is done by encoding an image into feature vectors for each patch, forming low-level features.
A separate network summarizes the low-level features into high-level features.
These low and high level features are grouped together but some high-level features are grouped with low-level features from another random image.
A discriminator is trained to assign correct groupings with a higher score than random groupings.

\citet{DBLP:journals/corr/abs-1807-03748} segments an image into overlapping patches, imposes a top-left to bottom-right ordering of all patches, then uses an auto-regressive model to predict ``future" patches of the image using InfoNCE loss.
``Future" is defined as the next patch in the imposed ordering.
Follow up work demonstrated that adding more model capacity and increasing the task difficulty (e.g., predicting several steps into the future) improves performance \citep{henaff2019data}.
\citet{DBLP:journals/corr/abs-1906-02940} proposes a similar approach but avoids imposing an ordering of patches by randomly masking input image patches and training the model to predict the masked patches.

\subsubsection{Channel Prediction}
Color or channel prediction methods perform self-supervised learning by removing channel information from an image and asking the model to predict the missing channel.
Several work \citep{DBLP:conf/eccv/ZhangIE16,DBLP:conf/cvpr/LarssonMS17,DBLP:conf/eccv/LarssonMS16} has shown that using colorization as a pre-training task can lead to improvements on ImageNet classification without labels.
In Split-Brain Autoencoders \citep{DBLP:conf/cvpr/ZhangIE17}, the authors split a traditional autoencoder into two disjoint sub-networks with each sub-network receiving a subset of the input channels.
The disjoint autoencoders are then trained to predict the missing channels of the other encoder.

\subsubsection{Temporal Prediction}
Temporal prediction focuses on exploiting temporal information to learn representations.
Many work in this area are based on ideas from early work on slow feature analysis (SFA) \citep{wiskott2002slow}, which suggests that a good prior for feature extraction is to learn features that vary slowly with time.
Learning to extract information that move slowly with time can naturally lead to higher-level representations and discard low-level noise.
A modern realization of this idea in deep learning is found in \citet{jayaraman2016slow}, where temporally close representations are encouraged to exhibit small differences.

Temporal prediction for computer vision focus on learning how to predict different perspectives of an image by leveraging either the camera movement or the motion of objects in the image.
A motivation for this type of learning is that motion in video helps identify objects, since pixels of the same moving object will likely move together \citep{wertheimer1938laws}.
In \citet{srivastava2015unsupervised}, the authors predict future frames in a video using an LSTM.
Alternatively, a contrastive loss can be used to avoid modeling low level information \citep{han2019video}.
Instead of learning directly to predict future frames, the motion of objects can be extracted as synthetic labels for training static images \citep{pathak2017learning}.

Temporal prediction has also been applied for self-supervised learning for speech.
\citet{schneider2019wav2vec} trains a self-supervised model from raw audio waves to predict future speech features against negative samples from the same audio clip, similar to \citet{DBLP:journals/corr/abs-1807-03748}.
\citet{chi2020audio} proposes to mask random speech frames (represented a spectrograms) and to predict those masked frames.
These approaches have many commonalities with those in section \ref{sec:spatial_pred}.



\subsubsection{Order Prediction}
Order prediction approaches aim to train a model to predict the position of image patches.
In \citet{doersch2015unsupervised}, random pairs of image patches are sampled from one of 8 positions in the image.
The model is asked to predict the relative position of one patch to another.
In \citet{noroozi2016unsupervised}, image patches are randomly shuffled and the model has to predict the permutation of the shuffle as a classification task.
A follow up work increased the difficulty \citep{kim2018learning} of the task by randomly deleting an image patch and asking the model to also predict the color of the image.
\citet{misra2016shuffle} applies this principle of order prediction to videos to predict the ordering of frames given shuffled frames.

\subsubsection{Hybrid Approaches}
When choosing a self-supervision task, it is not necessary for us to choose only a single predictive learning task. Recent work \citep{chen2020simple} has shown that a combination of different self-supervision tasks can yield much better results, rivaling the results of purely supervised learning for image classification.

\subsection{Pre-training for Discrete Domains}
In this section, we survey approaches that enable self-supervised learning in the discrete domain such as natural language processing (NLP).
Natural language treats text as a sequences of discrete symbols (also called tokens).
Although we primarily focus on self-supervised learning applied to NLP, techniques presented here are likely applicable to other forms of discrete sequences or non-natural languages (e.g.,~modeling music \citep{donahue2019lakhnes} or programming languages \citep{DBLP:journals/corr/abs-2006-03511}).

\subsubsection{Word Embeddings}
Skip-gram and Continuous Bag of Words (CBOW) \citep{DBLP:journals/corr/abs-1301-3781, DBLP:conf/nips/MikolovSCCD13} are popular approaches developed in 2013 for learning high quality word embeddings.
Skip-gram learns word embeddings by forcing words to predict nearby surrounding words within a given context.
Given a context of word embeddings $S = (s_{t-c},...,s_t,...,s_{t+c})$ with context length $c$, Skip-gram predicts $s_{t+i}, i \in [-c, c], i \neq t $ from $s_t$.

CBOW involves a similar idea to Skip-gram, but instead learns to predict $s_t$ using the sum of its surrounding embeddings,
\begin{equation}
\hat s_t = \sum_{i \in [-c, c], i \neq t} s_i.
\end{equation}
Once these embeddings are learned they can be used as input or fine-tuned as lower layers of other models.

\subsubsection{Contextual Embeddings}
Word embeddings are scalable and fast to train, but are limited in their representative power since they are usually learned using a linear model.
Furthermore, words in isolation provide limited information for which features can be extracted.
A natural extension to word embeddings is to learn deeper networks with contextual embeddings.

Early work explored learning contextual representations by predicting contiguous sentences of an input using a recurrent neural network \citep{kiros2015skip}.
Contextual Word Vectors \citep{mccann2017learned} provided embeddings based on a word and its entire sentence by leveraging the attention learned from machine translation.
These models have shown some success in text classification tasks and question answering.

\subsubsection{Language Models}
Core to the recent surge in transfer learning in NLP arises from the success of self-supervised learning from language modeling tasks and their variants.
Language modeling, in this context, is a pre-training task that learn to predict the probability of the next word or token given a historical context for an input sequence $X=\{x_1,...,x_n\}$.
\begin{equation}
    p(x_i | x_1,...,x_{i-1})
\end{equation}
A seminal work that demonstrates the general transferability of language modeling is the paper \textit{Embeddings from Language Models} (ELMo) \citep{DBLP:journals/access/SunYLWZLW19}. ELMo learns a bidirectional LSTM \citep{hochreiter1997long} language model and demonstrated strong improvements to a variety of downstream GLUE tasks with less labeled data.

\paragraph{Transformer Language Models}
Since ELMo, researchers have transitioned to focus on training self-attention models instead of recurrent neural networks.
Transformers \citep{DBLP:conf/nips/VaswaniSPUJGKP17} are a type of deep neural network that contain stacked layers of self-attention and feed-forward layers.
When compared to recurrent neural networks, Transformers are more efficient to train and enable gradients signals to easily propagate to all positions of the input.
The General Pre-trained Transformer (GPT) \citep{radford2018improving} is the first successful attempt at pre-training a Transformer and achieving strong target task performance for a variety of tasks.
GPT learns a unidirectional language model on a large corpus of text.
Follow up work \citep{brown2020language} scaled GPT to larger models and bigger datasets, observing strong generative capabilities and zero-shot performance on a variety of natural language tasks.

\paragraph{Masked Language Modeling}
A major limitation with GPT is that it learns a unidirectional language model in which every token can only attend to the tokens left of it.
Bidirectional Encoder Representations for Transformers (BERT) \citep{devlin2018bert,baevski2019cloze} proposes to learn bidirectional Transformers using a masked language modeling (MLM) pre-training task.
MLM randomly removes input tokens to the model and trains the model to predict the removed tokens.
At every iteration, BERT masks 15\% of its input tokens.
The downside of BERT is the pre-training procedure is expensive (only 15\% of positions are trained per iteration) and it does not explicitly learn conditional generation akin to language models.
Several extensions of BERT have been proposed, such as SpanBERT \citep{joshi2020spanbert}, a training procedure that masks out contiguous spans instead of individual tokens, and ERNIE \citep{sun2019ernie}, which masks out full entities or phrase-level units.
These strategies propose \textit{smarter} masking strategies for better performance.

\paragraph{Permutation Language Models}
XLNet \citep{yang2019xlnet} harnesses the benefits of language model conditioning with bidirectional training by introducing a permutation language modeling objective.
However, BERT, with more training and better hyperparameters, can outperform XLNet \citep{yang2019xlnet}.
It is later shown that permutation language modeling can be seen as a masked language model with stochastic attention masks \citep{kong2019mutual}.

\subsubsection{Sequence to Sequence Pre-training}
BERT has shown a lot of success in natural language inference tasks, but it is less well suited for sequence to sequence tasks.
Pre-training for sequence to sequence learning is explored T5 \citep{raffel2019exploring}, BART \citep{lewis2019bart} and MASS \citep{song2019mass}.
\citet{raffel2019exploring} provides an extensive analysis of various sequence to sequence pre-training tasks including prefix language modeling, masking and deshuffling.
They found that masking input spans and asking the model to generate these masked spans leads to the best performance.
Interestingly, learning how to deshuffle an input sequence performs the worse, which contradicts some of the success of order prediction techniques found in vision.

\subsubsection{Discriminative Pre-training Tasks}
An alternative to the popular approach of learning a generative language model is to consider discriminative pre-training tasks.
Indeed, in the original BERT \citep{devlin2018bert} implementation the authors proposed to jointly perform masked language modeling and next sentence prediction.
Next sentence prediction is a task where segments of text (specifically, sentences) are randomly swapped 50\% of the time and the model must predict whether or not the swap occurred.
This task has later been found to be not useful given masked language modeling  \citep{liu2019roberta}.

Electra \citep{clark2020electra} proposes to pre-train a model by classifying whether or not a token in the input sequence was randomly replaced by a small BERT model.
This focuses the model to learn how to differentiate real sequences from plausible alternatives.
The authors demonstrated that learning a discriminator yields strong results on downstream tasks with much better sample efficiency, since every single position is trained per iteration.




\section{Discussion}
\label{sec:frontiers}
Throughout this review, we have seen a variety of approaches to enable self-supervised learning.
The following are some general tips for self-supervision based on our observations of previous work.


\paragraph{Pre-training should be challenging}
Choosing a pre-training task that is sufficiently difficult is desired  and the difficulty should scale as models becomes larger.
It is also critical to prevent models from exploiting shortcuts and cheating \citep{minderer2020automatic} or leak statistical information from normalization \citep{chen2020simple}.
Combining pre-training tasks can be much better than using any single pre-training task alone \citep{chen2020simple}.
Ideally, pre-training tasks should be similar to the target task or subsume it.
For example, language modeling have shown to implicitly perform few shot learning when the dataset and model is sufficiently large \citep{brown2020language}, likely because the patterns that appear in a text corpus naturally contain relevant tasks.
Designing better and more universal\footnote{Universal can be defined as beneficial to all conceivable tasks that humans care about, which does not contradict the ``No Free Lunch Theorem" \citep{wolpert2012no}.} pre-training tasks should be an active area for future investigation.

\paragraph{More data and larger models are better}
Unsurprisingly, having more data and larger models lead to better results \citep{bigtransfer}. This is even more important in self-supervised learning where a lot of information needs to be absorbed for fine-tuning.
Furthermore, as seen in the trend of moving from word embeddings to contextual embeddings in NLP, the more parameters of a model that are pre-trained the better.
Even under computational constraints, training a larger model with more parameters for fewer iterations on a sufficiently large dataset is better than training a small model \citep{li2020train}.
These large models can be subsequently pruned if fast inference is required \citep{frankle2018lottery}.

\paragraph{Choose flexible architectures}
Choosing model architectures that have more flexibility (trading off priors and bias) can be advantageous in the context of self-supervised learning.
More flexible models enable a form of soft architectural search \citep{elsken2018neural}.
We see this example in NLP where Transformers have the advantage of having no positional bias as opposed to the receny bias of recurrent neural networks \citep{DBLP:conf/naacl/RavfogelGL19}.
This lack of positional bias likely provides more opportunities for gradient descent to mold its learning, which explains Transformer's tendency to be more data hungry and appropriate for large scale self-supervised training.
In computer vision, most self-supervised work has focused on ResNet \citep{he2015deep} and it would be interesting to see if this trend holds across domains.

\subsection{Future Work}
There are many future directions to further explore self-supervised learning.
Simply scaling existing approaches to larger models and datasets have diminishing returns \citep{bigtransfer} and even 175 billion parameter language models cannot learn commonsense physics and lack world knowledge \citep{brown2020language}.
Most self-supervised learning approaches have been focused on a single domain and it would be interesting to extend these techniques to multi-modal scenarios.
After all, humans are multi-modal learners.
Several work \citep{chen2019uniter,arandjelovic2017look} have shown promising results in this direction by performing contrastive learning of audio and visual information or masked ``language" modeling between images and text.

Another area to explore is better ways to extract information from these pre-trained models.
In this survey, we primarily focused on the popular fine-tuning approach, but other knowledge adaptation techniques exist \citep{ruder2019neural}.
For example, \citet{liu2019multi,raffel2019exploring} explored learning multiple tasks in a multi-task learning framework while fine-tuning pre-trained language models, leading to better downstream performances.
One interesting approach for pre-trained language models adopted by \citet{brown2020language} is few-shot \textit{probing}.
This technique involves using natural language itself to specify the desired downstream task along with a few examples and requires no fine-tuning.
It would be interesting to see if this type of \textit{probing} works for other domains such as vision and speech.


\section{Conclusion}
Supervised learning's primary bottleneck is the availability of labeled data.
Self-supervised learning is a powerful technique to extract knowledge from a large unlabelled corpus of data.
After a model is trained in a self-supervised manner, it can attain significantly improved performance on tasks that have few labels and even on tasks that have plenty of labels. 
The value of self-supervision comes from its scalability with virtually unlimited data in certain domains and its ability to be fine-tuned to a variety of tasks.
In the long term, self-supervision approaches are likely to outperform more task-specific approaches as computational resources become more ubiquitous \citep{sutton2019bitter,lecun2015deep}.

\subsubsection*{Acknowledgments}
Thanks to Professor Garrison W. Cottrell for providing comments, advice and editorial assistance.
Thanks to Bodhisattwa Prasad Majumder for providing proofreading assistance.

\bibliography{refs}

\end{document}